\documentclass[letterpaper,10pt,journal,twoside]{IEEEtran}
\usepackage{amsmath,amsfonts}
\usepackage{algorithmic}
\usepackage{algorithm}
\usepackage{array}
\usepackage{multirow}
\usepackage[caption=false,font=normalsize,labelfont=sf,textfont=sf]{subfig}
\usepackage{textcomp}
\usepackage{array}
\usepackage{stfloats}
\usepackage{url}
\usepackage{verbatim}
\usepackage{graphicx}
\usepackage{booktabs}
\usepackage{caption}
\usepackage{cite}
\usepackage{stfloats}
\usepackage{hyperref}
\usepackage{amsmath,amsfonts}
\usepackage{algorithmic}
\usepackage{algorithm}
\usepackage{array}
\usepackage[caption=false,font=normalsize,labelfont=sf,textfont=sf]{subfig}
\usepackage{textcomp}
\usepackage{stfloats}
\usepackage{url}
\usepackage{verbatim}
\usepackage{graphicx}
\usepackage{bm}
\usepackage{tabularx}
\usepackage[dvipsnames]{xcolor}
\usepackage{hyperref}
\usepackage{makecell}
\usepackage[inkscapelatex=false]{svg}
\usepackage{pifont}
\newcommand{\cmark}

\usepackage{diagbox}
\usepackage{makecell}
\usepackage{enumitem}

\newcommand{\mbf}[1]{\mathbf{#1}}
\DeclareRobustCommand{\revised}[1]{\textcolor{black}{#1}}

\begin{document}
\bstctlcite{IEEEexample:BSTcontrol} 

\title{REFINE-DP: Diffusion Policy Fine-tuning for Humanoid Loco-manipulation via Reinforcement Learning}
\author{Zhaoyuan Gu*, Yipu Chen*, Zimeng Chai*, Alfred Cueva, Thong Nguyen$^{\dagger}$, Yifan Wu$^{\dagger}$, Huishu Xue$^{\dagger}$, \\Minji Kim, Isaac Legene, Fukang Liu, KyoungMok Kim, Ayan Barula, Yongxin Chen, Ye Zhao
\thanks{\hspace*{1em}Manuscript received: March 1, 2026; Revised: May 31, 2026; Accepted: July 1, 2026.
\par\hspace*{1em}This paper was recommended for publication by Editor Olivier Stasse upon evaluation of the Associate Editor and Reviewers' comments.
\par\hspace*{1em}This work was supported in part by the Office of Naval Research (ONR) under Grant N000142312223, in part by the National Science Foundation (NSF) under Grant CMMI-2144309, Grant 2328254, and Grant 2409016, and in part by the United States Department of Agriculture (USDA) under Grant 2022-11065.
\par\hspace*{1em}The authors are with the Institute for Robotics and Intelligent Machines, Georgia Institute of Technology, Atlanta, GA 30332 USA (e-mail: \{zgu78, yezhao\}@gatech.edu). (* $^{\dagger}$ equally contributed)
\par\hspace*{1em}Digital Object Identifier (DOI): see top of this page.}}

\markboth{IEEE Robotics and Automation Letters. Preprint Version. Accepted July, 2026}
{Gu \MakeLowercase{\textit{et al.}}: REFINE-DP: Diffusion Policy Fine-tuning for Humanoid Loco-manipulation via Reinforcement Learning}
\maketitle

\begin{abstract}
Humanoid loco-manipulation requires coordinated task-space motion planning with stable loco-manipulation command tracking under complex robot-environment dynamics and long-horizon tasks. While diffusion policies (DPs) show promise for learning from demonstrations, deploying them on humanoids poses critical challenges: the motion planner trained offline is decoupled from the loco-manipulation controller, leading to poor command tracking, compounding distribution shift, and task failures. The common approach of scaling demonstration data is prohibitively expensive for high-dimensional humanoid systems. To address this challenge, we present REFINE-DP (REinforcement learning FINE-tuning of Diffusion Policy), a hierarchical framework that jointly optimizes a DP motion planner and an RL-based loco-manipulation controller. The DP is fine-tuned via a PPO-based diffusion policy gradient to improve task success rate, while the controller is simultaneously updated to accurately track the planner's evolving command distribution, reducing the distributional mismatch that degrades motion quality. We validate REFINE-DP on a humanoid robot performing loco-manipulation tasks, including door traversal and long-horizon object transport. REFINE-DP achieves an over $90\%$ success rate in simulation, even in out-of-distribution cases not seen in the pre-training data, and enables real-world execution without privileged state information. Our proposed method substantially outperforms pre-trained DP baselines and demonstrates that RL fine-tuning is key to reliable humanoid loco-manipulation. \href{https://refine-dp.github.io/REFINE-DP/}{https://refine-dp.github.io/REFINE-DP/}
\end{abstract}


\section{Introduction}

\IEEEPARstart{H}{umanoid} robots are increasingly capable in their physical behaviors, yet they still lack the task-level autonomy required to perform human-like tasks, such as coordinating locomotion and manipulation for door traversal. Recent studies have demonstrated agile whole-body control with both model-based methods~\cite{khazoom2024tailoring, jang2026seec} and learning-based motion imitation~\cite{he2024omnih2o, humanplus}. Despite these advances, existing approaches often struggle to operate autonomously in dynamic environments, with reliable execution still depending significantly on human supervision or heuristic planning.
To perform physical tasks reliably, humanoid robots must interact with the physical world through contact, such as manipulating objects and exerting force on the environment, where modeling errors and contact uncertainties frequently lead to task failure. These challenges require real-time adaptation beyond motion tracking control.

To address these challenges, recent work has explored learning task-level action generation directly from demonstrations, enabling robots to acquire task-level autonomy beyond predefined control policies.
For instance, recent advances in diffusion policies (DPs)~\cite{diffusion_policy} have introduced a promising behavior cloning (BC) approach for learning action-generative models from expert demonstrations. These models are effective at capturing complex and multimodal expert behaviors. However, their success often relies on offline datasets, which can lead to poor out-of-distribution robustness due to distribution shift and compounding errors.
To mitigate this issue, DPs are typically trained on large datasets that require high-capacity models, such as transformers, to improve coverage and generalization.
However, for high-dimensional humanoid systems, this scaling strategy incurs prohibitive costs in both data collection and policy training computation, while still failing to achieve robust performance in loco-manipulation, where long horizons, high dimensionality, and compounding execution errors exacerbate distribution mismatch.

Rather than scaling data, this paper addresses the distribution-shift issue through reinforcement learning fine-tuning (RLFT)~\cite{ren_diffusion_2024, jung2025ppf}, enabling adaptation beyond offline pre-training.
Specifically, we fine-tune a DP by exploring interactions in a simulator and collecting trials to directly update the DP via policy gradient~\cite{ren_diffusion_2024}. By exploring unseen state–action pairs, RLFT significantly increases the task success rate with only a sparse reward. Additionally, we demonstrate that RLFT enables a pre-trained DP on a small dataset to outperform those trained on substantially larger offline datasets.

\begin{figure*}[t]
\centerline{\includegraphics[width=0.7\textwidth]{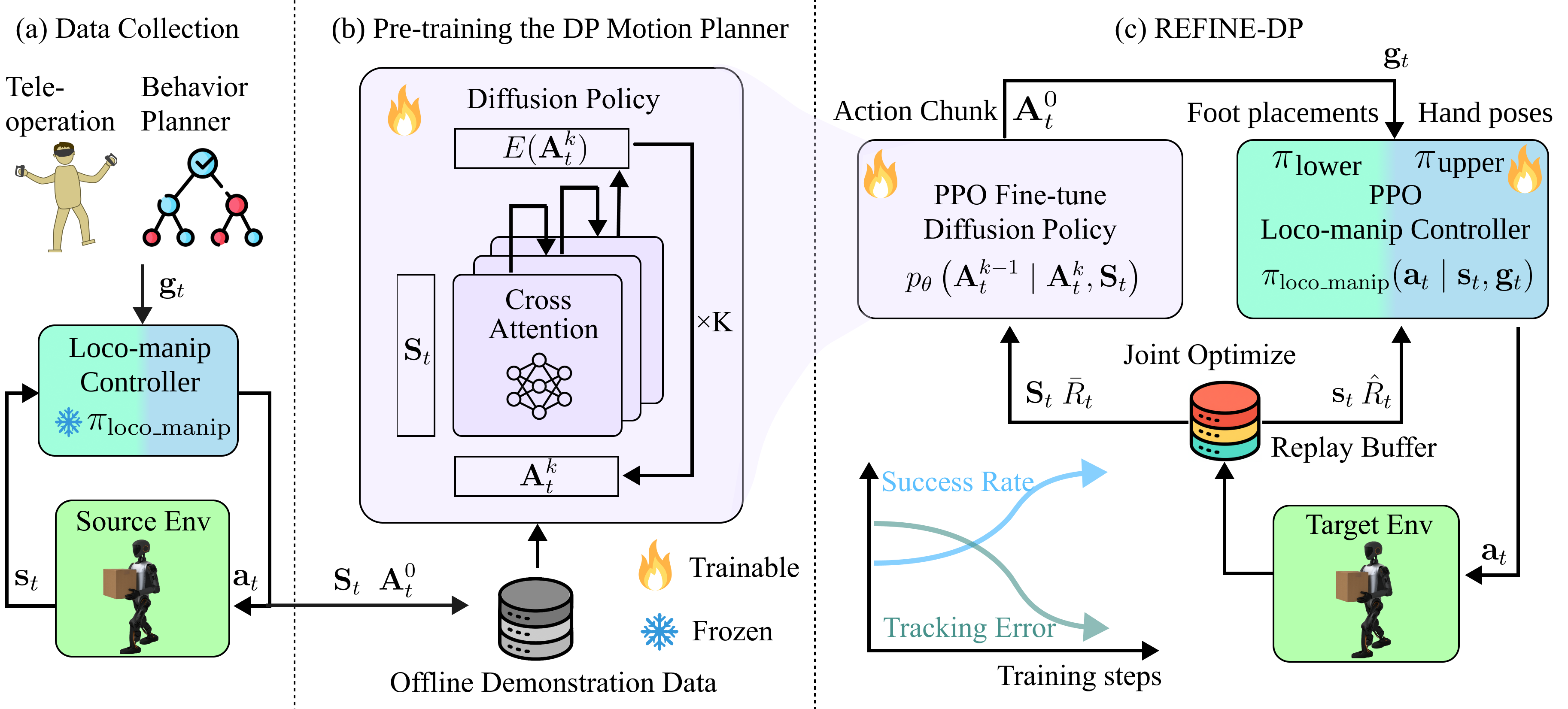}}
\caption{Our pipeline consists of three stages: (a) Data Collection, where expert demonstrations are collected in a source environment using a frozen RL-based loco-manipulation controller; (b) Pre-training, where a diffusion policy (DP) motion planner is trained with the expert dataset containing human-demonstrated skills; and (c) Joint Optimization, where the pre-trained DP is jointly fine-tuned with the loco-manipulation controller in a target environment. During deployment, the DP generates Cartesian action chunks, and the loco-manipulation controller converts them into joint-position references for tracking.}
\label{fig:framework}
\vspace{-0.2in}
\end{figure*}

Deploying DP on a humanoid robot remains a challenge due to the complexity of whole-body control.
To simplify the DP action space, we adopt a hierarchical framework with explicit roles for each layer. The DP serves as a motion planner that generates Cartesian action chunks, including base velocity and $SE(3)$ hand-pose trajectories. RL-based loco-manipulation controllers consume these action chunks and convert them into joint-position references while maintaining locomotion stability and manipulation accuracy. This hierarchical framework keeps the DP command space compact and intuitive while delegating the complexity of whole-body control to the dedicated loco-manipulation controller. An illustration of the proposed hierarchical framework is in Fig.~\ref{fig:framework}.

Based on this hierarchical framework, we introduce REFINE-DP (\textbf{RE}inforcement learning \textbf{FINE}-tuning of \textbf{D}iffusion \textbf{P}olicy), a novel fine-tuning approach that jointly optimizes both the DP motion planner and the RL-based loco-manipulation controller.
During joint optimization, the parameters of both the DP and the loco-manipulation controller are updated.
This joint optimization maintains distributional consistency between the planner's outputs and the controller's inputs, resulting in improved command-tracking performance and higher task success rates.

In summary, our main contributions are as follows:
\begin{itemize}[leftmargin=*]

\item We design a hierarchical humanoid loco-manipulation framework in which a DP serves as the motion planner and an RL policy acts as the loco-manipulation controller. Instead of operating directly in full-body configuration spaces, our DP outputs low-dimensional Cartesian action chunks (i.e., base velocities and $SE(3)$ hand-pose trajectories) tailored to humanoid loco-manipulation. These action chunks substantially simplify the DP action space, and can be easily teleoperated by human operators and converted by the RL policy into joint-position references.

\item We introduce a joint fine-tuning scheme that simultaneously optimizes a DP motion planner and an RL loco-manipulation controller. This is in contrast to prior RLFT schemes that update a single policy. We show that this joint optimization improves both task success rate and motion-tracking accuracy compared to fine-tuning either component alone.

\item We validate our framework on the T1 humanoid robot, enabling loco-manipulation across a range of tasks, including walking and door opening, long-horizon box transport, and stepping onto an elevated platform to retrieve an object. We further demonstrate the whole pipeline using only onboard RGB information for object pose estimation.
\end{itemize}

\vspace{-0.1in}
\section{Related Works}
\subsection{Sim-to-Real RL for Humanoid Loco-manipulation Control}

Loco-manipulation skills involve simultaneous locomotion and manipulation, posing a high-dimensional control problem for humanoid robots~\cite{gu_humanoid_2025}. Prior work has addressed this challenge using sim-to-real reinforcement learning (RL), such as for whole-body control frameworks~\cite{ze2025twist2}, while other approaches decompose the problem into upper- and lower-body controllers~\cite{zhang2025falcon, lu2024MobileTV}. Many of these RL methods~\cite{humanplus, he2024omnih2o, ze2025twist2, ben2025homie} support teleoperation, enabling demonstration data collection for training a motion planner.

Among those RL methods, the first category is learning from scratch, where the desired behavior is specified through a complex reward function. However, learning loco-manipulation without demonstrations is challenging due to the high dimensionality of humanoid systems and the diversity and compositional complexity of loco-manipulation tasks. Several studies have demonstrated successful sim-to-real transfer for autonomous object manipulation~\cite{zhang2024wococo, Curiosity-schwarke23a, SR_deepmind_soccer, xue2025openingsimtorealdoorhumanoid}. These policies are typically tuned for each specific task, making it difficult to acquire long-horizon, autonomous loco-manipulation skills.

The second category uses RL for motion tracking of reference, such as a manually designed trajectory \cite{dao2023simtoreal, liu2024opt2skill} or human data \cite{xie_box_locomanip, he2024omnih2o, humanplus}. While this paradigm has achieved versatile motion tracking capability, it fundamentally relies on externally provided references and therefore lacks the task- and environment-level awareness required for true autonomy~\cite{he2024omnih2o}.

In this paper, our loco-manipulation controller falls in the second category. It can walk and manipulate objects with desired base-velocity and hand-pose commands. Compared with other whole-body controllers with latent vector inputs~\cite{lu2024MobileTV} or whole-body joint-angle references~\cite{ze2025twist2}, our control policy's input space is more physically interpretable and enables more intuitive task-level control, either through VR teleoperation or a motion planner.

\subsection{Autonomous Humanoid Loco-Manipulation}

To achieve autonomous task execution on humanoid robots, many studies adopt modern behavior cloning (BC) techniques, such as diffusion policy (DP)~\cite{diffusion_policy} or action-chunking transformer~\cite{aloha}, to learn contact-rich manipulation directly from multi-modal demonstration data. However, while most autonomous policies focus on relatively stable manipulator platforms, extending these approaches to autonomous humanoid loco-manipulation remains challenging due to underactuated and unstable dynamics, and the need to coordinate whole-body motion over long horizons.

Among the few prior works on autonomous humanoid loco-manipulation, some approaches adopt hierarchical frameworks that decouple motion planning from loco-manipulation command tracking, thereby keeping the action space simple and interpretable. For example, \cite{ze2025twist2, humanplus, yin2025visualmimic} employ a planner–controller hierarchy to structure decision making and execution. Other approaches maintain a unified end-to-end policy but rely on large-scale domain randomization to integrate perception and improve sim-to-real robustness, as demonstrated in recent work on humanoid door opening~\cite{xue2025openingsimtorealdoorhumanoid, weng2025hdmi}. Our REFINE-DP's hierarchical framework falls into the first category, with a motion planner offering the advantage of learning a diverse range of tasks. Instead of freezing the control policy during the optimization of a planner policy \cite{sferrazza_humanoidbench_2024}, 
we jointly optimize both policies to simultaneously achieve better task success rate and tracking performance.

\subsection{Fine-tuning Pre-trained Imitation Learning Policy}

To address the inherent issue of compounding error and domain shift in DP, several works have explored integrating RL with DP by learning a residual policy on top of a pre-trained DP to refine actions and enhance performance. These residual policies provide corrective compensations for imitation errors. Existing works have explored on-policy RL~\cite{ankile2024imitationrefinementresidual, su2026rfsreinforcementlearningresidual} and off-policy RL~\cite{ankile2025residualoffpolicyrlfinetuning} for learning residual refinement to improve precision and sample efficiency.

Another line of work focuses on fine-tuning diffusion policies with RL. In Diffusion Policy Policy Optimization (DPPO)~\cite{ren_diffusion_2024}, the authors adapt Proximal Policy Optimization (PPO)~\cite{PPO} and fine-tune a pre-trained diffusion policy for a higher success rate. Another work \cite{ma2025efficientonlinereinforcementlearning} proposes two algorithms, Diffusion Policy Mirror Descent (DPMD) and Soft Diffusion Actor Critic (SDAC), for online training of diffusion policies from scratch. Both of these methods update the diffusion policies directly.
More broadly, $\pi_{RL}$~\cite{chen2026piRL} adopts RL for fine-tuning vision-language-action models.
While prior work optimizes only the planner, few works jointly optimize a diffusion-based motion planner and a loco-manipulation controller within a unified planning and control framework.

\section{Methods}

Our pipeline consists of three stages, as illustrated in Fig.~\ref{fig:framework}. It first collects expert loco-manipulation demonstrations, then pre-trains a diffusion policy (DP), which is subsequently fine-tuned to enhance the task success rate and motion quality.

\textbf{Collecting loco-manipulation data.}
Expert demonstrations are collected by teleoperating the robot or by rolling out a heuristic planner in IsaacLab~\cite{mittal2025isaaclab}. During the teleoperation, the operator gives commands via a VR device to a pre-trained loco-manipulation policy $\pi_{\text{loco\_manip}}$, as detailed in Sec.~\ref{sec:RL_train}. In addition to teleoperation, we leverage heuristic planners to scale up data collection. During data collection, a humanoid robot executes diverse loco-manipulation skills, such as object transportation and door opening.

\textbf{Pre-training the diffusion policy:}
For each task, we pre-train a DP $\bar{\pi}_{\theta}$ using the collected dataset. The pre-trained DP outputs base velocity and hand pose commands, which are passed to the $\pi_{\text{loco\_manip}}$ for execution.

\textbf{Fine-tuning in target environments:}
The pre-trained DP is fine-tuned in the simulator for the same or a more challenging target environment. We adopt two fine-tuning settings: (i) Only the diffusion policy parameters are updated, while the control policy $\pi_{\text{loco\_manip}}$ remains frozen as a loco-manipulation controller; (ii)
Jointly optimize both the $\bar{\pi}_{\theta}$ and $\pi_{\text{loco\_manip}}$.
Fine-tuning the DP adapts it to domain-specific dynamics beyond the offline expert data, substantially improving its success rate. Joint optimization further yields better motion quality, namely, more tracking accuracy and less motion jitter.

\subsection{Training Loco-manipulation Controller for Data Collection}
\label{sec:RL_train}

\begin{figure}[t]
  \centering
  \includegraphics[width=0.8\columnwidth]{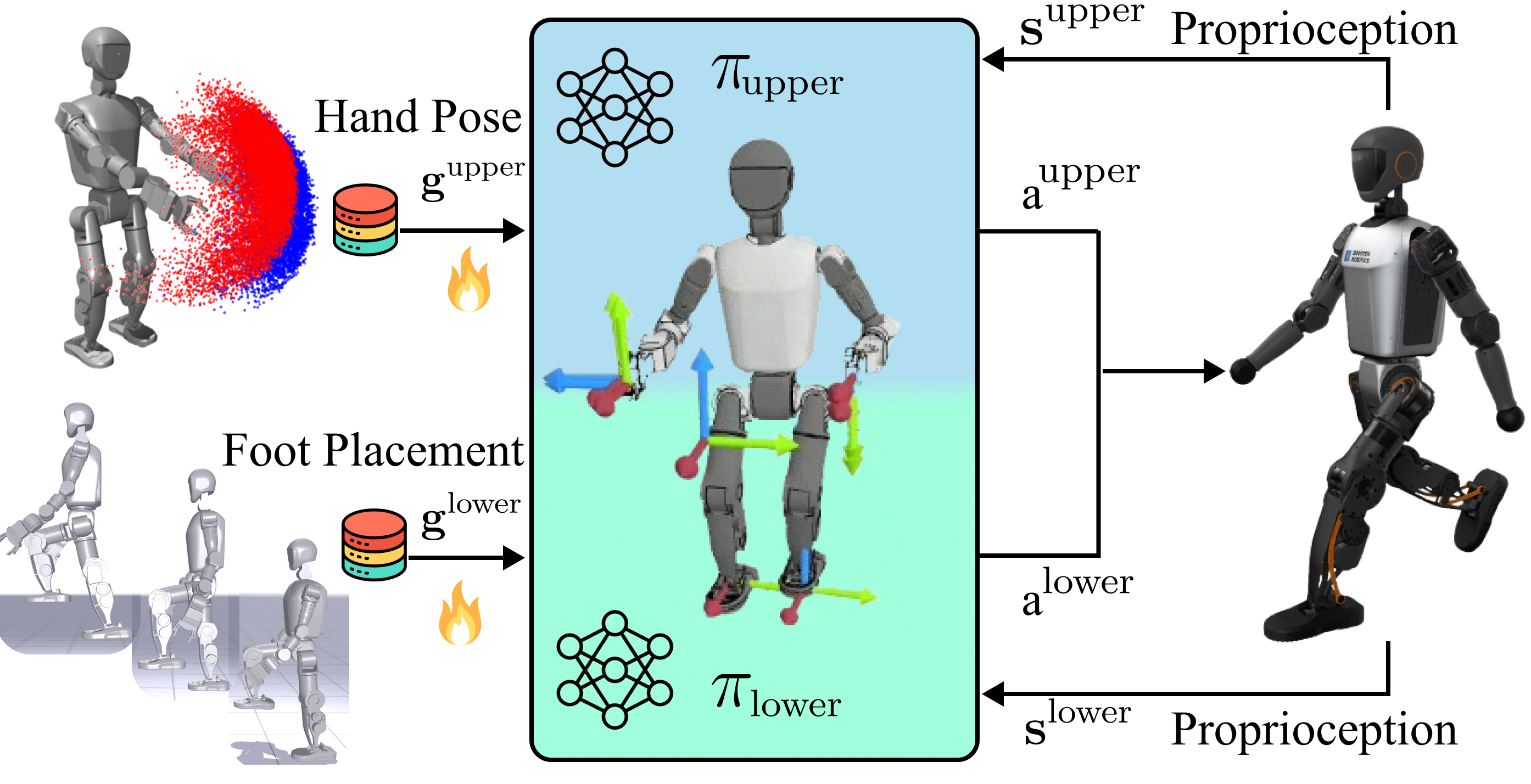}
\caption{The sim-to-real RL of the loco-manipulation control policy, from motion prior generation to deployment.}
  \label{fig:RL_flow}
\vspace{-0.25in}
\end{figure}

We develop a reinforcement learning (RL) policy $\pi_{\text{loco\_manip}}$ capable of simultaneous locomotion and manipulation.
The training and deployment pipeline is illustrated in Fig.~\ref{fig:RL_flow}. $\pi_{\text{loco\_manip}}$ adopts a decoupled control architecture that separates upper-body and lower-body behaviors, similar to prior humanoid loco-manipulation works~\cite{zhang2025falcon, ben2025homie, lu2024MobileTV}. Specifically, a lower-body locomotion policy is responsible for achieving intermediate foot placements while maintaining dynamic balance, whereas an upper-body arm policy tracks desired hand poses to execute manipulation tasks. During training, the two controllers are decoupled, handling disturbances from each other through domain randomization.

While most existing locomotion policies are formulated as velocity-tracking controllers, their objectives are designed for long-distance periodic walking rather than for the frequent start–stop transitions and precise torso position adjustments required for manipulation tasks. To enable accurate locomotion toward target positions and achieve a high loco-manipulation success rate, we introduce a foot-placement tracking controller $\pi_{\rm lower}$ that takes discrete foot-placement commands as input. Unlike velocity-tracking controllers that can accumulate positional error over time, our locomotion policy provides direct control over where each footstep lands. Prior studies~\cite{wang2025beamdojo, suliman2025reinforcement} have demonstrated that this formulation yields substantially improved foot-placement accuracy and locomotion stability.

We train $\pi_{\rm lower}$ using RL motion imitation~\cite{DeepMimic2018} on lower-body reference trajectories~\cite{liu2024opt2skill}. The policy $\pi_{\rm lower}(\mathbf{a}^{\rm lower}_t \mid \mathbf{s}^{\rm lower}_t, \mathbf{g}^{\rm lower}_t)$ is conditioned on a foot-placement command $\mathbf{g}^{\rm lower}_t$, which consists of a swing-foot indicator, a countdown, and a relative target swing-foot pose in the stance-foot frame. The proprioceptive observation $\mathbf{s}^{\rm lower}_t$ includes the joint states, the gravity vector, and the previous action. The reward function combines foot tracking terms with regularization terms to ensure accuracy and stability.
Additionally, we address disturbances from upper-body movement by training with randomized arm configurations.

The upper-body policy $\pi_{\rm upper}(\mbf{a}^{\rm upper}_t \mid \mbf{s}^{\rm upper}_t, \mbf{g}^{\rm upper}_t)$ tracks the hand-pose command $\mbf{g}^{\rm upper}_t$, which specifies $SE(3)$ targets for the left and right hands in the robot's root frame. We generate a set of feasible hand commands via sampling collision-free joint configurations and solving forward kinematics. The proprioceptive observation $\mbf{s}_t^{\rm upper}$ includes arm joint states, current hand poses, and tracking errors. To improve robustness to payload and external forces arising from object interaction and lower-body motion, we apply domain randomization during the training of $\pi_{\mathrm{upper}}$.

Both $\pi_{\mathrm{lower}}$ and $\pi_{\mathrm{upper}}$ output joint position commands $\mbf{a}_t = [\mbf{a}^{\mathrm{lower}}_t;\mbf{a}^{\mathrm{upper}}_t]$, where $\mbf{a}^{\mathrm{lower}}_t \in \mathbb{R}^{12}$ corresponds to the leg joints and $\mbf{a}^{\mathrm{upper}}_t \in \mathbb{R}^{14}$ corresponds to the arm joints. These commands specify joint position offsets relative to a fixed default configuration $\mbf{q}_{\mathrm{def}}$. The combined loco-manipulation policy is $\pi_{\text{loco\_manip}}(\mbf{a}_t \mid \mbf{s}_t, \mbf{g}_t)$, where $\mbf{s}_t = [\mbf{s}^{\mathrm{lower}}_t;\mbf{s}^{\mathrm{upper}}_t]$ is the proprioceptive observation and $\mbf{g}_t = [\mbf{g}^{\mathrm{lower}}_t;\mbf{g}^{\mathrm{upper}}_t]$ is the loco-manipulation command.
The resulting target joint positions, $\mbf{q}_{\mathrm{target}} = \mbf{q}_{\mathrm{def}} + \mbf{a}_t$, are tracked by proportional-derivative (PD) control. Detailed observations and rewards are on \href{https://refine-dp.github.io/REFINE-DP/}{our website}.

The primary methods for data collection include teleoperation and the rollout of heuristic planners, both leveraging $\pi_{\text{loco\_manip}}$ for loco-manipulation in a simulator. These heuristic planners are designed for each task via stage-conditioned behavior transitions, whose intermediate waypoints are parameterized based on observed object poses. Teleoperation and heuristic rollouts play complementary roles: teleoperation captures recovery behaviors but is costly and low-throughput because humanoid loco-manipulation requires whole-body coordination, whereas heuristic planners scale data collection at the cost of behavioral novelty. Considering this trade-off, we augment $50$ teleoperated trajectories with heuristic planner rollouts, expanding the dataset to $1000$ trajectories; the teleoperated trajectories define the core behavioral modes, including recovery responses, and heuristic rollouts augment coverage. To ensure broad coverage of scenarios, we randomize both the initial object configurations and the robot’s initial states during data collection; additional data diversity arises from locomotion drifting, which randomizes torso pose and manipulation timing. We collect only successful trajectories, each containing state-action pairs. The observed states include the robot's hand and foot poses in the body frame, the gripper state, and object information. The action specifies the desired hand poses, gripper state, and a base-velocity command. We use a velocity-to-footstep planner to convert the velocity command into a sequence of footstep commands.

\revised{Another choice of command interface is a pair of global hand poses. Although they are low-dimensional, they under-specify whole-body motion in a kinematically redundant humanoid and can yield undesirable torso configurations~\cite{nai2026humi}. Our interface instead combines hand-pose commands with a lower-body command, similar to OmniH2O and SONIC~\cite{he2024omnih2o,luo2025sonic}; this disambiguates locomotion from hand pose tracking and preserves explicit foothold control on rough terrain.}

\subsection{Diffusion Policy Pre-training}
The collected loco-manipulation data is used to pre-train a DP \cite{diffusion_policy}.
Formally, a DP models a conditional action distribution $p_\theta(\mbf{A}_t^{0:K} \mid \mbf{S}_t)
=
p(\mbf{A}_t^{K})
\prod_{k=1}^{K} p_\theta(\mbf{A}_t^{k-1} \mid \mbf{A}_t^{k}, \mbf{S}_t)$, where $\mbf{A}^k_t$ represents an action chunk, and $\mbf{S}_t$ represents a state observation chunk~\cite{aloha}. The superscript $k$ is the denoising timestep, $K$ is the total number of denoising steps, and the subscript $t$ is the environment timestep.
Given the dataset $\mathcal{D}=\{(\mbf{S}_i, \mbf{A}_i^0)\}_{i=1}^N$, we maximize the approximate log likelihood, which can be reformulated as training a noise prediction network $\epsilon_\theta$ by minimizing the noise prediction error \cite{diffusion_model}:
\begin{equation}
    \mathcal{L}_{\text{diff}}(\theta) = \mathbb{E}_{\mbf{A}_t^0, \mbf{S}_t, k, \epsilon}\Big[\|\epsilon - \epsilon_\theta(\mbf{A}_t^{k}, \mbf{S}_t, k)\|^2\Big],
\end{equation}
where $\epsilon\sim\mathcal{N}(0,\mathbf{I})$ is the target noise, and $\mbf{A}_t^k$ is $\mbf{A}_t^0$ corrupted by $\epsilon$ through a forward diffusion process.
The action generation involves a reverse denoising process.

\vspace{-0.1in}
\subsection{Diffusion Policy Fine-tuning}
\label{sec:dpft}
While using the DP as a motion planner achieves a decent success rate, its performance remains insufficient for reliable task execution. In particular, the pre-trained planner does not explicitly account for the robot’s closed-loop dynamics and execution errors, resulting in deviations that accumulate over long horizons and occasional task failure. This compounding error in pre-training motivates us to further enhance the policy. In this study, we employ RL to fine-tune the pre-trained DP. RL fine-tuning offers two key advantages: (i) improving task success through trial-and-error interaction with the simulation environment, and (ii)  enhancing robustness to out-of-distribution conditions through domain randomization, enabling zero-shot sim-to-real transfer.

Conventional policy gradient methods such as Proximal Policy Optimization (PPO)~\cite{PPO} rely on evaluating a policy density function $\bar{\pi}_\theta(\mbf{A}_t^0 |\mbf{S}_t)$.
However, DPs are implicit policies, for which $\bar{\pi}_\theta(\mbf{A}_t^0 |\mbf{S}_t)$
is not tractable. To address this, we adopt Diffusion Policy Policy Optimization (DPPO)~\cite{ren_diffusion_2024} that augments an environment Markov Decision Process (MDP) by incorporating the denoising process into the MDP and treating each denoising step as a decision step. Since denoising transitions follow a tractable Gaussian distribution, this augmented MDP enables a likelihood-based policy gradient method, such as PPO~\cite{PPO}, allowing RL fine-tuning of DP.

Formally, we define the environment MDP as $\mathcal{M}_{\text{ENV}} := ({\mathcal{S}, \mathcal{A}, P_0, P, R})$, where $\mathcal{S}$ is the state space, $\mathcal{A}$ is the action space, $P_0$ is the initial state distribution, $P$ is the transition probabilities, and $R$ is the reward function.
We then define a diffusion-process-augmented MDP $\mathcal{M}_\text{DP}:= ({\bar{\mathcal{S}}, \bar{\mathcal{A}}, \bar{P}_0, \bar{P}, \bar{R}})$,
which expands $\mathcal{M}_{\text{ENV}}$ by inserting one full denoising process in each environment timestep. This augmented $\mathcal{M}_\text{DP}$ defines a unified timestep index $\bar{t}(t,k) = tK + (K-1-k)$, where $K$ is the total number of denoising steps and $k\in[0, \ldots,K-1]$. Accordingly, the augmented MDP is given by
$$
\bar{s}_{\bar{t}(t,k)}=(\mbf{S}_t, \mbf{A}_t^{k+1}), \quad \bar{a}_{\bar{t}(t,k)}=\mbf{A}_t^k , \quad \bar{P}_0=P_0\otimes\mathcal{N}(0, \mathbf{I}).
$$
\begin{multline}
\bar{P}\!\left(\bar{s}_{\bar{t}+1} \mid \bar{s}_{\bar{t}}, \bar{a}_{\bar{t}}\right) = \!\!\\
\left\{
    \begin{array}{ll}
    (\mbf{S}_t, \mbf{A}_t^{k}) \sim \bm{\delta}_{\mbf{S}_t,\mbf{A}_t^k}, & k>0 \\[6pt]
    (\mbf{S}_{t+1}, \mbf{A}_{t+1}^{K}) \sim P(\mbf{S}_{t+1} \mid \mbf{S}_{t}, \mbf{A}_{t}^{0}) \otimes \mathcal{N}(0, \mathbf{I}), & k=0
    \end{array}
\right.
\raisetag{12pt}
\end{multline}
\begin{multline}
\bar{R}_{\bar{t}(t, k)}\left(\bar{s}_{\bar{t}(t, k)}, \bar{a}_{\bar{t}(t, k)}\right)=
\left\{
    \begin{array}{ll}
    0, & k>0 \\
    R_t(\mbf{S}_{t}, \mbf{A}_{t}^{0}), & k=0
    \end{array}
\right.
\end{multline}
where $\bm{\delta}$ is the Dirac distribution. We refer the readers to \cite{ren_diffusion_2024} for detailed derivation.

We use generalized advantage estimation (GAE) \cite{schulman_high-dimensional_2018} to calculate the advantage estimate $\hat{A}(\bar{s}_{\bar{t}},\bar{a}_{\bar{t}})$ from a replay buffer and optimize the DP using a PPO-style policy gradient.

\subsection{Joint Optimization of the Diffusion and the RL Policies}

To improve the motion quality beyond the success rate, we propose a joint optimization technique that fine-tunes both the DP motion planner and the RL-based loco-manipulation controller. Unlike the fine-tuning described in Sec.~\ref{sec:dpft} that optimizes only the DP, we additionally optimize the loco-manipulation control policy.

As shown in Fig.~\ref{fig:framework}(c), the augmented environment $\mathcal{M}_\text{DP}$ produces two sets of states $(\bar{s}_{{t}},\mbf{s}_{{t}})$ and rewards $(\bar{R}_{{t}}, \hat{R}_{{t}})$. $\bar{R}_{{t}}$ is used to update the DP $\bar{\pi}_{\theta}(\bar{a}_{{t}} \mid \bar{s}_{{t}})$ and enhance task success rate. $\hat{R}_{{t}}$ represents a set of rewards used in pre-training the loco-manipulation control policy $\pi_\textrm{loco\_manip}$, and its purpose is to further improve motion quality and enable accurate and smooth tracking of the DP commands.
Both the DPs and the RL control policy are optimized using PPO. A detailed description of the algorithm is in Alg.~\ref{alg:co-train}.

Joint optimization enables the tracking controller to accurately follow dynamic, task-relevant commands generated by the DP planner. Unlike the independently sampled, stationary commands used during controller pre-training, DP commands represent a moving target along a continuous trajectory, creating a distribution mismatch that degrades tracking performance. Joint optimization alleviates this mismatch by exposing the controller to planner-generated commands, bringing them in-distribution and substantially improving tracking accuracy. Combined with the optimization of DP, this process enhances motion quality and achieves a high success rate.

\begin{algorithm}[tb]
\caption{REFINE-DP: Joint Optimization of DP and RL}
\label{alg:co-train}
\begin{algorithmic}
\STATE {\bfseries Input:} Pre-trained diffusion policy $\bar{\pi}_\theta(\mbf{A}_t^0 |\mbf{S}_t)$, loco-manipulation policy $\pi_\textrm{loco\_manip}$, augmented MDP $\mathcal{M}_\text{DP}$.
\STATE Initialize state $\bar{s}_{{t}}$ from $\mathcal{M}_\text{DP}$.
\FOR{$iter=0$ {\bfseries to} $L$}
   \STATE Rollout to collect replay buffer $D$.
   \FOR{$iter=0$ {\bfseries to} $M$}
   \STATE Sample a mini-batch $D_k$ from $D$.
   \STATE Update $\pi_\textrm{loco\_manip}$ using $D_k$ via PPO~\cite{PPO}.
   \ENDFOR
   \STATE Rollout to collect replay buffer $\bar{D}$.
   \FOR{$iter=0$ {\bfseries to} $N$}
   \STATE Sample a mini-batch $\bar{D}_k$ from $\bar{D}$.
   \STATE Update $\bar{\pi}_\theta(\mbf{A}_t^0 |\mbf{S}_t)$ using $\bar{D}_k$ via DPPO. 
   \ENDFOR
\ENDFOR
\end{algorithmic}
\end{algorithm}

\section{Experiments}

\subsection{Experiment Setup}
\label{sec:experiment}

We use IsaacLab~\cite{mittal2025isaaclab} for expert demonstration collection, RL policy training, and fine-tuning. All policies are trained on an NVIDIA H200 GPU. RL pre-training of $\pi_{\rm loco\_manip}$ takes $10$h with $4096$ parallel environments, while DP pre-training on a $100$-trajectory dataset ($3$h of demonstrations) requires $18$h. The pipeline from pre-training to the fully fine-tuned policy requires $22$h of wall-clock time. Joint optimization (Alg.~\ref{alg:co-train}) takes $L=2$ iterations, each costing $9$h, split equally between updating $\pi_\textrm{loco\_manip}$ and $\bar{\pi}_\theta$.

Regarding policy configuration, the DP $\bar{\pi}_{\theta}$ has an observation horizon of $8$ and an action chunk size of $12$, with a $0.1$ s interval between each observation and action. We choose $\bar{\alpha}_k$ to follow the cosine schedule \cite{nichol_improved_2021} for DP pre-training.

For hardware experiments, we use a Booster T1 robot with 29 degrees of freedom. Both the DP motion planner $\bar{\pi}_{\theta}$ and the RL-based loco-manipulation controller $\pi_{\text{loco\_manip}}$ run on a policy computer, with an AMD 7945HX CPU and an NVIDIA RTX 4060 GPU. The loco-manipulation policy $\pi_{\rm loco\_manip}$ runs at $50$ Hz, while the DP $\bar{\pi}_{\theta}$ runs at $10$ Hz using NVIDIA TensorRT. This policy computer receives the robot’s proprioceptive states (i.e., joint angles and IMU measurements) from T1’s embedded computer. The policy computer also performs the inference of $\bar{\pi}_{\theta}$ that outputs a Cartesian action chunk for the $\pi_{\text{loco\_manip}}$, which then outputs upper- and lower-body joint-position references.
We use a state-based DP that relies on the pre-processed pose of the target object relative to the robot torso, which is captured using either a motion capture (MoCap) system or an onboard RGB camera. The MoCap system streams object-state observations at $90$ Hz. For the camera setting, a head-mounted Intel RealSense D435i provides video for 6-DoF fiducial-marker-based~\cite{olson2011apriltag} object-pose estimation at 30 Hz, demonstrating the capability to operate without privileged state information.

\begin{figure*}[t]
\centering
\includegraphics[width=0.65\textwidth]{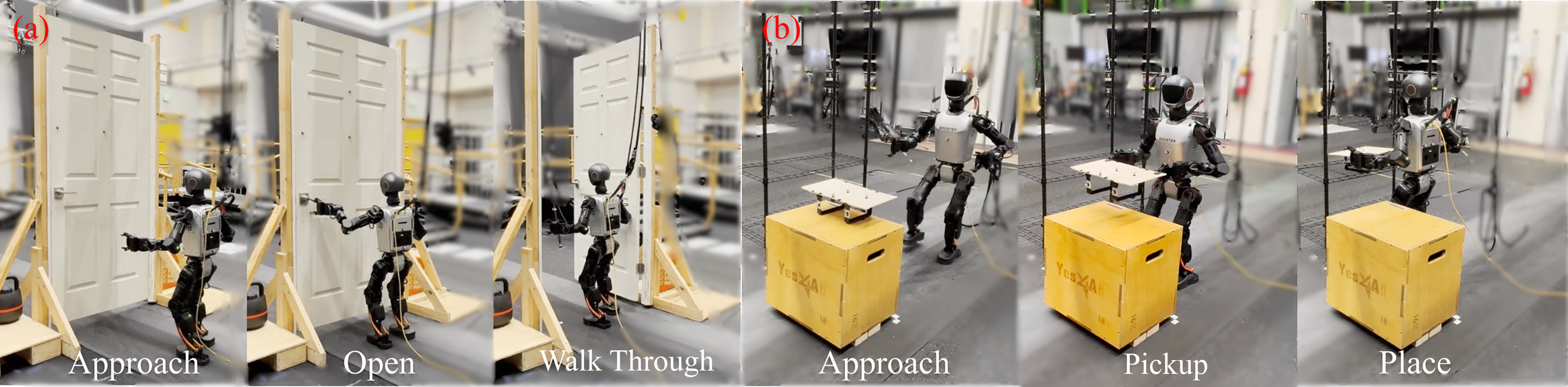}
\caption{The robot autonomously executes loco-manipulation tasks:
(a) door opening, (b) object transportation.}
\label{fig:hardware}
\vspace{-0.2in}
\end{figure*}

As shown in Fig.~\ref{fig:hardware}, we evaluate our REFINE-DP on a set of loco-manipulation tasks. Each of these tasks requires multiple stages of walking and manipulation. Unlike~\cite{StageACT_Lee2025_door}, which requires a stage signal, our DP policy infers task stages from object locations and implicitly learns stage transitions, enabling loco-manipulation without explicit stage supervision.

\textbf{Task 1: Object pickup; Task 2: Long-horizon pick-and-place.}
Task 1 requires walking to a table to pick up a box, whereas Task 2 is a longer-horizon extension (i.e., $40$ s) with placing the object on another table.

\textbf{Task 3: Door opening and traversal.}
For humanoid door-opening tasks, prior work relies on heuristic planners~\cite{calvert2024behaviorarchitecturefasthumanoid}, sim-to-real RL~\cite{ xue2025openingsimtorealdoorhumanoid}, or action-chunking transformer (ACT)~\cite{StageACT_Lee2025_door}. In contrast, we are the first to apply DP to humanoid door traversal.

\textbf{Task 4: Stair-assisted object retrieval.}
This loco-manipulation task involves uneven terrain, where the humanoid must step onto an elevated platform to retrieve an object.

\subsection{Baseline Methods and Ablation Study}
\label{sec:baseline}
\revised{We compare REFINE-DP with planner baselines spanning architectures and fine-tuning strategies: pre-trained DiT, LSTM, and deterministic MLP planners; RL-fine-tuned MLP (MLP-FT); and residual RL~\cite{ankile2024imitationrefinementresidual}. All use the same frozen loco-manipulation controller $\pi_{\text{loco\_manip}}$.}

\textbf{DiT.}
The DP baseline follows the standard formulation~\cite{diffusion_policy} using a transformer backbone.
This evaluates the performance of a purely pre-trained DP within our hierarchical framework.

\textbf{LSTM.} The LSTM baseline serves as a recurrent baseline to the transformer-based diffusion policy (DiT).  


\textbf{MLP} and \textbf{MLP-FT.}
The MLP planner is a separate baseline under the same planner-controller interface, rather than a replacement module inside the learned denoising diffusion process.
It directly regresses actions from observations and therefore produces deterministic action predictions.
In contrast, the DiT-based diffusion planner samples action trajectories through a denoising diffusion process, which allows stochastic trajectory generation.
\revised{MLP-FT retains the hierarchical interface: only the MLP planner is fine-tuned, while the deterministic low-level controller remains frozen. This matches DiT-FT and isolates the contribution of the diffusion-based policy formulation from the fine-tuning procedure itself.}
Since PPO-style fine-tuning relies on stochastic exploration, we convert the pre-trained deterministic MLP into a stochastic policy by interpreting its output as the mean action and sampling around it with an Ornstein Uhlenbeck (OU) process.
Specifically, the OU process is a mean-reverting diffusion process, producing temporally correlated and smoothly varying perturbations that are suitable for continuous-control exploration:
$$
\mbf{A}_t^{k-1} = (1 - \lambda_k)\mbf{A}_t^{k}
+ \lambda_k \tilde{\bm{\mu}}_\theta(\mbf{S}_t)
+ \tilde{\bm{\sigma}}_k \bm{\epsilon}_k,
\quad \bm{\epsilon}_k \sim \mathcal{N}(0, \mathbf{I}),
$$
where $\tilde{\bm{\mu}}_\theta(\mbf{S}_t)$ denotes the MLP mean action prediction.
The interpolation coefficient $\lambda_k$ pulls samples toward the mean policy, while $\tilde{\bm{\sigma}}_k$ controls the exploration magnitude.
We adopt a linear schedule for $\lambda_k$ and a decreasing noise schedule for $\tilde{\bm{\sigma}}_k$, encouraging exploration in early steps and concentration around the mean policy in later steps. 
\revised{Compared with the OU process that injects noise to the planner, exploration using $\pi_{\text{loco\_manip}}$'s standard deviation instead produced excessive joint-level noise, preventing task completion and yielding no success signal.}

\textbf{Residual RL.}
Residual RL~\cite{ankile2024imitationrefinementresidual} improves a pre-trained DP by learning a lightweight corrective residual that is added to the output of the frozen DP. Rather than modifying the parameters of a pre-trained DP, a residual Gaussian policy is optimized with PPO to predict small additive adjustments to the frozen DP's actions.
\begin{figure}[t]\centering\includegraphics[width=0.45\textwidth]{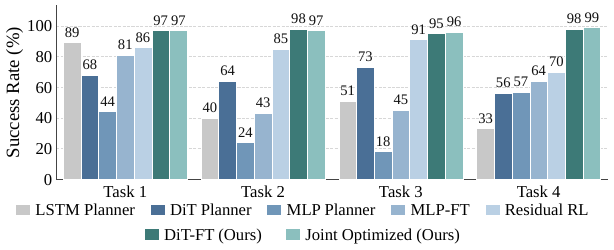}
  \caption{Comparison of success rates across tasks in Sec.~\ref{sec:experiment} and methods in Sec.~\ref{sec:baseline}.}
  \label{fig:sr_table}
  \vspace{-0.2in}
\end{figure}

\subsection{Quantitative Results and Analysis}
\label{sec:result}
\subsection*{C-1 Task Success and Planner Capacity}

\textbf{REFINE-DP improves task success rate (SR).}
The task SR comparison with the baselines is reported in Fig.~\ref{fig:sr_table}. The results show that REFINE-DP, including fine-tuning and joint optimization, substantially outperforms all baselines across all loco-manipulation tasks. REFINE-DP can achieve more than $90\%$ by fine-tuning from the pre-trained policy of $50-70\%$. Since the fine-tuned SR is already high, joint optimization does not further improve SR; instead, it improves motion quality.

\textbf{Stochastic policy shows a superior SR.}
We observe a consistent performance gain in the transformer-based diffusion (DiT) over the MLP-based diffusion in Fig.~\ref{fig:sr_table}.
We attribute this improvement to DiT's ability to model multi-modal action distributions and generate diverse feasible trajectories. In contrast, MLPs perform deterministic regression and tend to average across demonstrations. When multiple valid solutions exist, this averaging effect can produce actions that may not correspond to a viable trajectory.

MLP-FT partially mitigates this limitation by injecting stochasticity during fine-tuning, enabling exploration and recovery from averaged behaviors. Performance improves as the policy adapts to the target environment and reduces the distribution gap between offline demonstrations and online execution.
However, the achievable improvement remains limited by the representational capacity of the backbone.
These results highlight the importance of backbone capacity for multi-modal action modeling and the effectiveness of fine-tuning.

LSTM can achieve comparable SR to DiT in the short-horizon Task 1.
However, it degrades sharply on long-horizon tasks, suggesting its recurrent state is insufficient to capture the action distribution. 

\textbf{Pure RL from sparse reward fails.} Successful fine-tuning requires a sufficiently capable pre-trained policy. We find that fine-tuning is effective when a pre-trained policy achieves a moderate success rate (approximately $50-70\%$). In contrast, fine-tuning from random initialization (Pure RL) shows limited improvement, as sparse rewards rarely provide informative learning signals when task success is infrequent. Pre-training alleviates the need for extensive reward shaping typically required in pure RL from scratch~\cite{xue2025openingsimtorealdoorhumanoid}. This allows the policy to improve its SR with minimal manual tuning.
\subsection*{C-2 Efficiency Gains from Fine-tuning and Joint Optimization}
\begin{figure}[t]
  \centering  \includegraphics[width=0.45\textwidth]{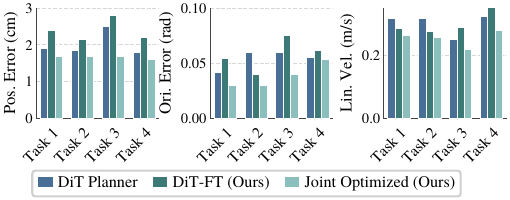}
  \caption{Joint optimization achieves better end-effector tracking performance and velocities across tasks in Sec.~\ref{sec:experiment}.}
  \label{fig:door_cotraining_vs_ft}
  \vspace{-0.2in}
\end{figure}

\textbf{Joint optimization improves fine-tuning efficiency and motion quality.} Optimizing the loco-manipulation controller provides three benefits. First, optimizing the RL controller alone using pre-trained DP rollouts $D$ already improves the SR by $18\%$ on the long-horizon pick-and-place task. Second, the jointly optimized loco-manipulation policy $\pi'_{\rm {loco\_manip}}$ improves the training efficiency of DP fine-tuning, requiring approximately half as many iterations ($20$ instead of $40$) to achieve a $90\%$ SR compared to fine-tuning with the pre-trained RL-based loco-manipulation policy $\pi_{\rm {loco\_manip}}$.
Third, joint optimization improves motion quality in addition to a consistently high SR, as shown in Fig.~\ref{fig:sr_table}. Specifically, we evaluate upper-body tracking performance using position and orientation errors, as well as motion smoothness using linear velocity, averaged across $100$ trials. Collectively, $\pi'_{\rm loco\_manip}$ achieves the lowest position error and reduces orientation error by up to $50\%$ compared to the pre-trained $\pi_{\rm {loco\_manip}}$, as shown in Fig.~\ref{fig:door_cotraining_vs_ft}. This strictly lower tracking error coincides with an average $15\%$ decrease in end-effector velocity, reflecting smoother manipulation. Meanwhile, fine-tuning the DP alone can degrade controller tracking performance, as the fine-tuned DP learns to issue excessive commands that reduce tracking accuracy, exert large forces, and produce jerky motion at unexpected contact changes.

\begin{figure}[t]
  \centering
  \includegraphics[width=0.45\textwidth]{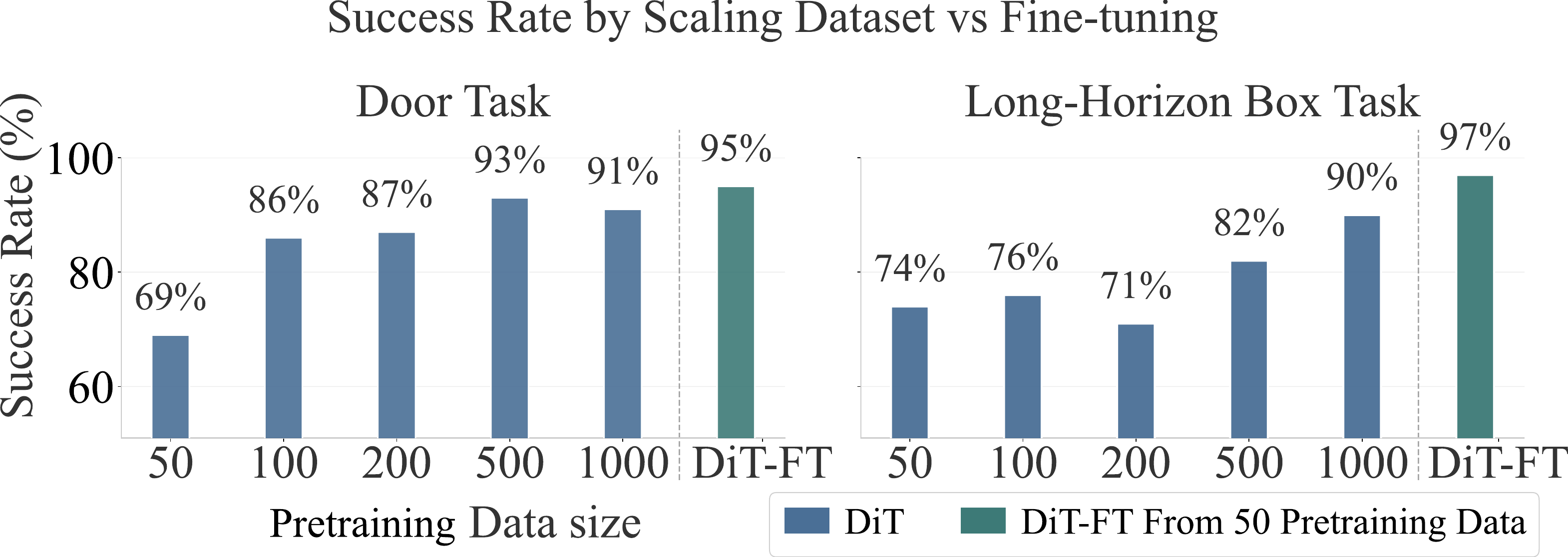}
  \caption{Comparison of success rate between scaling pre-training data and fine-tuning diffusion policy.}
  \label{fig:scale_vs_ft}
  \vspace{-0.2in}
\end{figure}

\textbf{Fine-tuning improves data efficiency. }
Fig.~\ref{fig:scale_vs_ft} compares the data scaling behavior with and without fine-tuning on both the box-pick-and-place and door-opening tasks.
While the SR improves with more pre-training data, fine-tuning achieves higher performance with substantially fewer demonstrations.
In particular, reaching a $90\%$ SR through pre-training alone requires approximately $1,000$ trajectories.
In contrast, a DP pre-trained on only $50$ trajectories achieves up to $95\%$ SR after fine-tuning. These results demonstrate that fine-tuning significantly reduces the need for expensive pre-training data while yielding a higher final SR.

\textbf{Fine-tuning adapts to out-of-distribution (OOD) scenarios in pre-training.} To extend the pre-trained policy to OOD scenarios, we employ domain randomization with a custom curriculum to expose it to scenarios beyond those observed during pre-training. Specifically, we fine-tune for the object transport task in unseen robot initial locations, which are parameterized by three variables relative to the target object: (i) radial distance, (ii) polar angle, and (iii) heading angle of the robot. The fine-tuning expands the ranges to cover $320\%$, $125\%$, and $600\%$ of the corresponding pre-training ranges, respectively.
While the pre-trained policy achieves only $0\%$ SR at the maximum randomization level, our curriculum-based fine-tuning improves the SR to over $80\%$. 


\subsection{Hardware Experiment}
We demonstrate autonomous loco-manipulation on a Booster T1 humanoid robot, as shown in Fig.~\ref{fig:hardware}. For stable and accurate execution, we clamp locomotion speed to $0.2$ \text{m/s} and hand speed to $0.05$ \text{m/s}. This enables accurate loco-manipulation, leading to higher SR.
In real-world experiments, REFINE-DP achieves a success rate of $70\%$ (Task 1), $50\%$ (Task 2), and $75\%$ (Task 3), respectively, over $N=20$ trials. The failure modes include occasional tripping of the lower-body controller and unsuccessful grasps on box or handles.

The sim-to-real gap arises from two primary sources. First, the object observations exhibit systematic offsets from object-pose calibration and become biased under MoCap or perception occlusions, especially during contact-rich phases such as grasping a box or door handle. Second, sim-to-real dynamics mismatch causes locomotion trembling and sliding, which increases the difficulty of manipulation. These factors lead to a reduction from the $90\%+$ SR in simulation.
Despite these gaps, the fine-tuned policy demonstrates improved robustness to noisy observations and execution errors.

REFINE-DP adapts to environmental perturbations during execution by re-adjusting and re-attempting. As demonstrated in Task 1, when the target object location is altered, the robot re-adjusts its plan on-the-fly and continues the task.
Upon execution failures, the policy re-attempts the task. For example, in the door-opening experiment, when the robot misses the door handle, it often takes small footsteps to get closer before attempting the manipulation again. Such corrective behaviors occur consistently under the fine-tuned policy, whereas the pre-trained policy often becomes stuck and fails to recover, leading to halted execution.
This capability to re-attempt and recover from unexpected changes and failed actions demonstrates the robustness and autonomy of our REFINE-DP.

During fine-tuning, REFINE-DP improves task efficiency by eliminating redundant and indecisive motions commonly observed in pre-trained policies, thereby increasing task throughput. REFINE-DP achieves average speedups of around $10\%$ and $20\%$ for the box pickup and door-opening tasks.



\section{Conclusion}

In this study, we introduced REFINE-DP, a fine-tuning framework that jointly optimizes a hierarchical planner–controller architecture. Fine-tuning the DP motion planner improves data efficiency and enhances generalization to out-of-distribution scenarios. Meanwhile, fine-tuning the loco-manipulation controller improves tracking accuracy and execution smoothness. Together, these components achieve high success rates on loco-manipulation tasks on a humanoid robot. A current limitation is that the planner remains state-conditioned, which requires an explicit object pose. Future work will focus on an end-to-end RGB-conditioned diffusion-policy planner, more versatile skills in a unified whole-body controller, and a richer planner-controller interface. Pre-training or fine-tuning with real-world data will be explored to better close the sim-to-real gap. In addition, this study relies on VR teleoperation and per-task heuristic rollouts; supplementing these with more scalable data sources, such as egocentric human data, is a promising direction for reducing the data burden.

\bibliographystyle{IEEEtran}
\bibliography{references.bib}

\newpage

\section{Appendix}

\subsection{Foot-Placement Policy Training}

As described in Sec.~\ref{sec:RL_train}, the foot-placement policy $\pi_{\rm lower}$ is conditioned on discrete foot-placement commands $\mathbf{g}^{\rm lower}_t$ with three components: (1) a swing-foot indicator specifying which foot is active, (2) a normalized phase countdown progressing from $1$ to $0$, and (3) a target swing-foot pose relative to the stance foot, parameterized by $(x,y,z,\mathrm{yaw})$. Given $\mathbf{g}^{\rm lower}_t$, $\pi_{\rm lower}$ generates a stepping motion that places the swing foot at the commanded target pose. During training, the policy learns this behavior by tracking a set of reference trajectories. During inference, however, it receives only the foot-placement command $\mathbf{g}^{\rm lower}_t$ and does not require access to the reference trajectories.

The trajectories are generated using a linear inverted pendulum model (LIPM) combined with lower-body inverse kinematics. Each reference motion corresponds to a specific stair rise and tread depth and consists of three phases: pre-swing, swing, and post-swing. To increase motion diversity, we randomize the lift-off and landing positions of the swing foot relative to the stance foot using discretized grid sampling. Given the sampled lift-off and landing positions, the corresponding swing-foot trajectory is precomputed as a cubic spline that avoids collisions with the stairs. We then solve a trajectory optimization problem to produce a joint-level stepping motion.

\begin{figure}[t]
\centering
\includegraphics[width=\columnwidth]{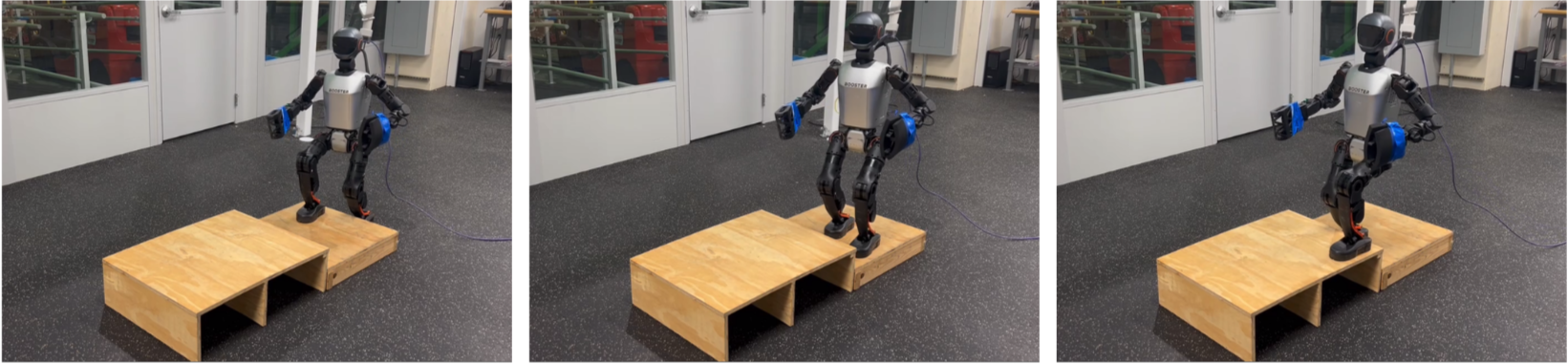}
\caption{The foot-placement policy enables the robot to walk on flat ground and stairs.}
\label{fig:stair}
\vspace{-0.15in}
\end{figure}

During training, the robot learns to track the reference stepping motions while maintaining balance on stairs. In each parallel simulation environment, the robot is initialized on a stair step whose rise-and-tread configuration matches the pre-swing and post-swing poses associated with the sampled reference trajectory. Detailed observations are listed in Table~\ref{tab:loco_manip_obs}. Table~\ref{tab:loco_rewards} summarizes the reward terms used during training.

Driven by these inputs and motion-imitation objectives, the policy learns to execute the targeted swing-foot placements precisely. Fig.~\ref{fig:stair} illustrates the resulting behavior as the robot navigates a multi-level platform. 

\subsection{Foot-Placement and Velocity-Tracking Ablation}

The loco-manipulation tasks in this work require precise torso positioning in addition to hand tracking because small base-pose errors can prevent the hands from reaching objects such as door handles, boxes, or stair platforms. Velocity-tracking interfaces do not directly enforce this positional objective. In practice, low-speed velocity commands can lead to stepping in place, overshoot, or unresponsiveness to small commands when a dead zone is used. In contrast, our foot-placement command specifies where each step should land, giving the controller direct control over the number and locations of footsteps used to approach the target. This explicit command interface is therefore better suited to the frequent stopping, restarting, and fine positioning required for humanoid loco-manipulation.

\begin{figure}[t]
\centering
\includegraphics[width=\columnwidth]{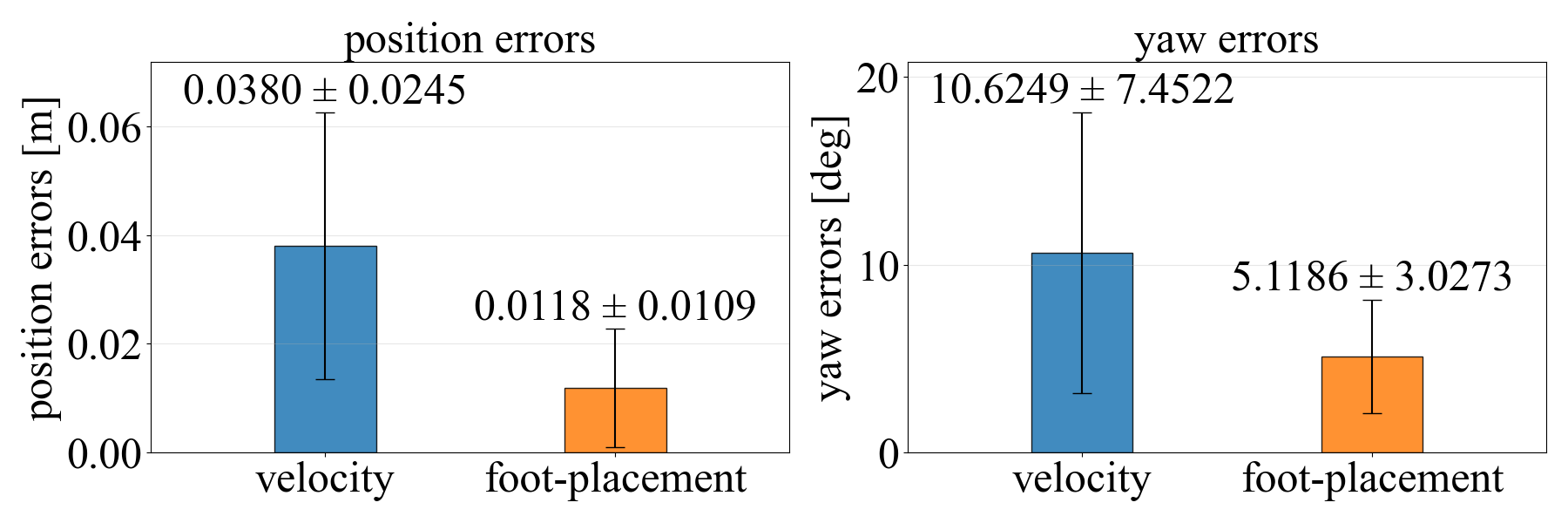}
\caption{Position and yaw tracking errors across 10 trials for velocity-based and foot-placement-based lower-body policies.}
\label{fig:vel_stair_plot}
\vspace{-0.05in}
\end{figure}

To quantify this effect, we conducted a torso-position-tracking ablation that isolates the lower-body command interface. We kept all upper-body components fixed and used the same PD feedback controller for torso-position tracking while varying only the lower-body policy. For the foot-placement policy, the feedback command is converted into explicit footstep targets by the velocity-to-footstep planner described in Sec.~\ref{sec:RL_train}. For the velocity-tracking policy, the same feedback command is used directly as a velocity command. We then measured the resulting torso position and yaw errors as the robot attempted to reach the target position. As shown in Fig.~\ref{fig:vel_stair_plot}, the foot-placement policy achieves lower tracking error and more consistent target reaching across trials. These results support our choice of foot-placement tracking over velocity tracking for precise loco-manipulation.
\begin{table}[t]
\centering
\caption{Observation terms for RL policy training}
\label{tab:loco_manip_obs}
\small
\begin{tabular}{l c c}
\hline
\textbf{Observation} & \textbf{Actor} & \textbf{Critic} \\
\hline
\multicolumn{3}{l}{\textit{Manipulation}} \\
\hline
Projected gravity & \checkmark & \checkmark \\
Arm joint pos. & \checkmark & \checkmark \\
Arm joint vel. & \checkmark & \checkmark \\
Action & \checkmark & \checkmark \\
Hand pose cmd. ($\mathbf{p}_L^{\text{cmd}}, \mathbf{p}_R^{\text{cmd}}$) & \checkmark & \checkmark \\
Current hand poses ($\mathbf{p}_L, \mathbf{p}_R$) & \checkmark & \checkmark \\
Hand tracking error ($\mathbf{e}_L, \mathbf{e}_R$) & \checkmark & \checkmark \\
\hline
\multicolumn{3}{l}{\textit{Locomotion}} \\
\hline
Foot placement cmd. & \checkmark & \checkmark \\
Base angular vel. & \checkmark & \checkmark \\
Projected gravity & \checkmark & \checkmark \\
Leg joint pos. & \checkmark & \checkmark \\
Leg joint vel. & \checkmark & \checkmark \\
Action & \checkmark & \checkmark \\
Reference foot pos. & & \checkmark \\
Current foot pos. & & \checkmark \\
Base linear vel. & & \checkmark \\
Foot wrench & & \checkmark \\
\hline
\end{tabular}
\end{table}

\begin{table}[t]
\caption{Reward terms for foot-placement tracking policy}
\label{tab:loco_rewards}
\centering
\small
\setlength{\tabcolsep}{3pt}
\renewcommand{\arraystretch}{1.05}
\begin{tabularx}{\columnwidth}{@{}l >{\raggedright\arraybackslash}X c@{}}
\toprule
\textbf{Term} & \textbf{Formulation} & \textbf{Weight} \\
\midrule
\multicolumn{3}{l}{\textit{End-effector reference tracking (per foot)}} \\
Position ($\sigma=0.05$) &
$\exp(-\lVert\mathbf{p}-\mathbf{p}^{\mathrm{ref}}\rVert^{2}/\sigma^{2})$ & $+5.0$ \\
Yaw ($\sigma=0.1$) &
$\exp(-(\psi-\psi^{\mathrm{ref}})^{2}/\sigma^{2})$ & $+3.0$ \\
Base vel. ($\sigma=0.25$) &
$\exp(-\lVert\mathbf{v}-\mathbf{v}^{\mathrm{ref}}\rVert^{2}/\sigma^{2})$ & $+1.0$ \\
\midrule
\multicolumn{3}{l}{\textit{Joint reference tracking}} \\
Position ($\sigma=0.3$) &
$\exp(-\lVert\mathbf{q}-\mathbf{q}^{\mathrm{ref}}\rVert^{2}/\sigma^{2})$ & $+0.5$ \\
Velocity ($\sigma=1.0$) &
$\exp(-\lVert\dot{\mathbf{q}}-\dot{\mathbf{q}}^{\mathrm{ref}}\rVert^{2}/\sigma^{2})$ & $+0.3$ \\
\midrule
\multicolumn{3}{l}{\textit{Regularization (with curriculum $\rightarrow$)}} \\
Action smoothness &
$\lVert\mathbf{a}_t-\mathbf{a}_{t-1}\rVert^{2}$ & $-0.001 \rightarrow -0.5$ \\
Joint acceleration &
$\lVert\ddot{\mathbf{q}}\rVert^{2}$ & $-10^{-7} \rightarrow -10^{-5}$ \\
Torque limits &
$\max(0,\lvert\boldsymbol{\tau}\rvert-\boldsymbol{\tau}_{\max})$ & $-0.5$ \\
Joint limits &
$\max(0,\lvert\mathbf{q}\rvert-\mathbf{q}_{\max})$ & $-10^{-4}$ \\
Ankle posture &
$\lVert\mathbf{q}-\mathbf{q}_{\mathrm{def}}\rVert^{2}$ & $-2.0$ \\
Base ori. ($\sigma=0.2$) &
$\exp(-\lVert\mathbf{g}_{\mathrm{xy}}\rVert^{2}/\sigma^{2})$ & $+3.0$ \\
Foot airtime ($\tau=0.4$) &
$\sum_{i} c_i(t^{\mathrm{air}}_i-\tau)$ & $+20.0$ \\
Base height range &
$\begin{aligned}
&[\max(0,h_{\min}-h)]^2\\[-1pt]
&+[\max(0,h-h_{\max})]^2
\end{aligned}$ & $-3.0$ \\
\bottomrule
\end{tabularx}
\end{table}

\begin{table}[t]
\caption{Reward terms for hand-pose tracking policy}
\label{tab:manip_rewards}
\centering
\small
\setlength{\tabcolsep}{4pt}
\renewcommand{\arraystretch}{1.05}
\begin{tabularx}{\columnwidth}{@{}l >{\raggedright\arraybackslash}X c@{}}
\toprule
\textbf{Term} & \textbf{Formulation} & \textbf{Weight} \\
\midrule
\multicolumn{3}{l}{\textit{End-effector position tracking (per hand)}} \\
Coarse ($\sigma=0.4$)   & $\exp(-\lVert\mathbf{p}-\mathbf{p}^{\mathrm{cmd}}\rVert^2/\sigma^2)$ & $+2.0$ \\
Fine ($\sigma=0.1$)     & $\exp(-\lVert\mathbf{p}-\mathbf{p}^{\mathrm{cmd}}\rVert^2/\sigma^2)$ & $+2.0$ \\
Precise ($\sigma=0.08$) & $\exp(-\lVert\mathbf{p}-\mathbf{p}^{\mathrm{cmd}}\rVert^2/\sigma^2)$ & $+2.0$ \\
\midrule
\multicolumn{3}{l}{\textit{End-effector orientation tracking (per hand)}} \\
Coarse ($\sigma=0.8$)   & $\exp(-\lVert\mathbf{e}_{\mathrm{quat}}\rVert^2/\sigma^2)$ & $+1.0$ \\
Fine ($\sigma=0.5$)     & $\exp(-\lVert\mathbf{e}_{\mathrm{quat}}\rVert^2/\sigma^2)$ & $+1.0$ \\
Precise ($\sigma=0.3$)  & $\exp(-\lVert\mathbf{e}_{\mathrm{quat}}\rVert^2/\sigma^2)$ & $+1.0$ \\
\midrule
\multicolumn{3}{l}{\textit{Regularization (with curriculum $\rightarrow$)}} \\
Posture prior       & $\lVert\mathbf{q}-\mathbf{q}_{\mathrm{nom}}\rVert_1$ & $-0.2$ \\
Action smoothness   & $\lVert\mathbf{a}_t-\mathbf{a}_{t-1}\rVert^2$       & $-0.01 \rightarrow -0.1$ \\
Joint velocity      & $\lVert\dot{\mathbf{q}}\rVert^2$                      & $-10^{-3} \rightarrow -2\times 10^{-3}$ \\
Joint acceleration  & $\lVert\ddot{\mathbf{q}}\rVert^2$                     & $-10^{-6} \rightarrow -3\times 10^{-6}$\\
EE acceleration     & $\lVert\ddot{\mathbf{p}}_{\mathrm{ee}}\rVert^2$         & $-1.4\times 10^{-2}$ \\
Torque limits       & $\max(0,\lvert\boldsymbol{\tau}\rvert-\boldsymbol{\tau}_{\max})$ & $-0.1$ \\
Joint limits        & $\max(0,\lvert\mathbf{q}\rvert-\mathbf{q}_{\max})$   & $-4.0$ \\
\bottomrule
\end{tabularx}
\end{table}

\subsection{Additional Results}
\label{apx:ood}

\begin{figure}[t]
  \centering
  \includegraphics[width=0.50\textwidth]{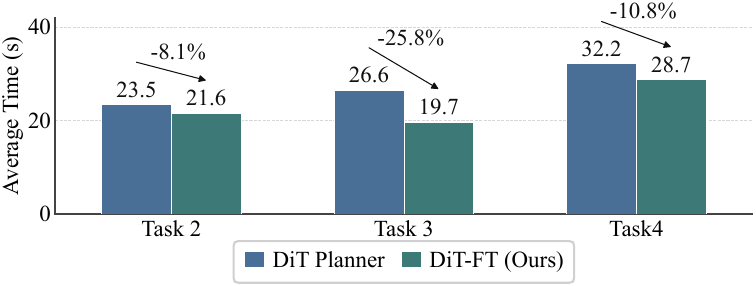}
  \caption{Fine-tuning reduces task execution time in both simulation and hardware experiments.}
  \label{fig:throughput}
  \vspace{-0.15in}
\end{figure}

\textbf{Fine-tuning achieves higher throughput.} The fine-tuned policy demonstrates higher throughput, completing the same task in less time. As shown in Fig.~\ref{fig:throughput}, fine-tuning reduces task completion time by $15\%$ on average across Task~2, Task~3, and Task~4, due to the improved action efficiency acquired through RL. In contrast, the pre-trained policy often produces hesitant or indecisive motions, resulting in slower execution.

\begin{figure}[t]
  \centering
    \centering
      \includegraphics[width=0.75\columnwidth]{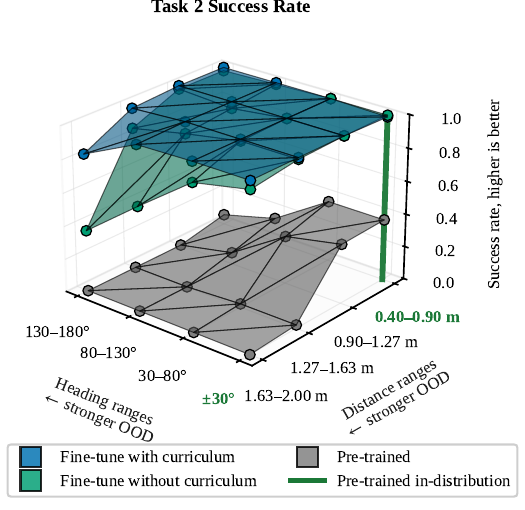}\\[-0pt]
      \vspace{-0.5em}
    {\small (a) Success rate}\\[5pt]
    \centering
    \includegraphics[width=0.75\columnwidth]{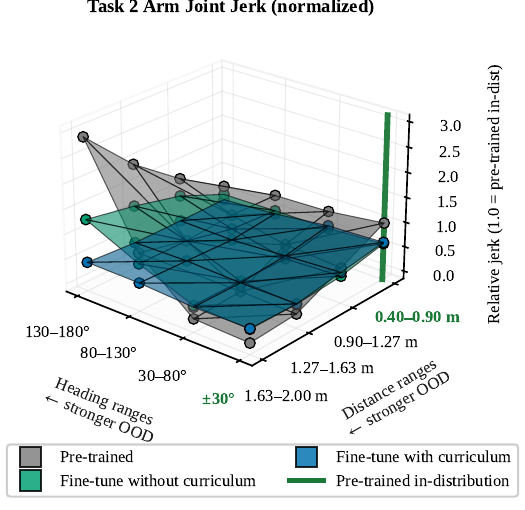}\\[-2pt]
    \vspace{-0.5em}
    {\small (b) Average jerk}
  \caption{(a) Evaluation success rates on different robot initialization ranges. With curriculum checkpoint shows the best performance in high OOD regions. (b) Average sum of upper body joint jerk squared over 100 rollout environments.}
  \label{fig:ood_sr}
  \vspace{-0.15in}
\end{figure}

\textbf{Fine-tuning using curriculum improves motion quality.} To improve the fine-tuning efficiency with high domain randomization, we introduce a curriculum that progressively increases the randomization ranges by $10\%$ once the policy achieves $90\%$ SR at the current randomization level. This curriculum enables the policy to gradually expand coverage and master increasingly challenging conditions. 
To evaluate the effectiveness of our proposed curriculum, we compare the final success rates of the pre-trained, curriculum-fine-tuned, and non-curriculum-fine-tuned checkpoints across different regions of the full randomization space, defined by the combination of heading and distance intervals. Both fine-tunes have been trained with the same hyper-parameters. The result in Fig.~\ref{fig:ood_sr} (a) shows that curriculum training adapt the desired behavior to OOD scenarios. In contrast, although fine-tuning without curriculum (directly training on the max randomization ranges) also leads to increased SR, the learned arm behavior exhibits more jitter in OOD scenarios due to the early exposure to high randomization scenarios, where the stochastic actions during rollout are amplified by PPO. This is illustrated by Fig.~\ref{fig:ood_sr} (b), where we report the squared sum of upper-body joint jerk averaged over 100 rollout environments. The with-curriculum fine-tune maintains stable jerk across all OOD and in-distribution (ID) scenarios, whereas the w/o-curriculum fine-tune shows much higher jerk, similar to the pre-trained checkpoint, when deployed to OOD settings.

\textbf{Adding noise to the DP state produces chaotic motion and lowers the success rate.}
{To test whether injecting observation noise improves the diffusion policy's (DP) robustness to the sim-to-real gap, we fine-tune DP planners under several observation-noise levels by adding zero-mean Gaussian noise to the observed state. As a result, noise with a standard deviation of $1$~cm and $3$~cm reduces the real-world Task~1 success rate from $70\%$ to $25\%$ and $20\%$, respectively. The dominant failure mode is excessive, chaotic motion in the predicted hand poses. We attribute this to the generative nature of the DP: perturbing the observed state corrupts the conditioning signal, so the planner learns to map noisy, inconsistent conditions to actions and produces unstable predictions rather than improved robustness.}

\end{document}